# Examining and Mitigating Kernel Saturation in Convolutional Neural Networks using Negative Images

Nidhi Gowdra, Roopak Sinha and Stephen MacDonell
*School of Engineering, Computer and Mathematical Sciences*
*Auckland University of Technology (AUT)*
*Auckland, New Zealand*
nidhi.gowdra@aut.ac.nz, roopak.sinha@aut.ac.nz, stephen.macdonell@aut.ac.nz

**Abstract**

*Neural saturation in Deep Neural Networks (DNNs) has been studied extensively, but remains relatively unexplored in Convolutional Neural Networks (CNNs). Understanding and alleviating the effects of convolutional kernel saturation is critical for enhancing CNN models classification accuracies. In this paper, we analyze the effect of convolutional kernel saturation in CNNs and propose a simple data augmentation technique to mitigate saturation and increase classification accuracy, by supplementing negative images to the training dataset. We hypothesize that greater semantic feature information can be extracted using negative images since they have the same structural information as standard images but differ in their data representations. Varied data representations decrease the probability of kernel saturation and thus increase the effectiveness of kernel weight updates. The two datasets selected to evaluate our hypothesis were CIFAR-10 and STL-10 as they have similar image classes but differ in image resolutions thus making for a better understanding of the saturation phenomenon. MNIST dataset was used to highlight the ineffectiveness of the technique for linearly separable data. The ResNet CNN architecture was chosen since the skip connections in the network ensure the most important features contributing the most to classification accuracy are retained. Our results show that CNNs are indeed susceptible to convolutional kernel saturation and that supplementing negative images to the training dataset can offer a statistically significant increase in classification accuracies when compared against models trained on the original datasets. Our results present accuracy increases of 6.98% and 3.16% on the STL-10 and CIFAR-10 datasets respectively.*

## 1. INTRODUCTION

Convolutional Neural Networks (CNNs) are the state-of-the-art for complex computer vision tasks such as image classification, image localization [1], [2], speech recognition [3] and natural language processing [4]. The two facets of a CNN model are the convolutional and classification blocks. The convolutional block extracts features from the input data and generates feature maps that are passed onto the classification block for final recognition and generation of class probabilities. The classification block is a conventional multi-layer Artificial Neural Network (ANN) with multiple layers of fully connected neurons. Failure to achieve optimal feature extractions in the convolutional block leads to knock-on effects in the classification block, causing suboptimal classification performance. While an all-convolutional network has been proposed by [5], state-of-the-art CNN models such as Residual Network (ResNet) [6] still use a traditional setup with a final fully connected classification block.

While state-of-the-art models such as ResNet-50 perform extremely well, they are susceptible to training afflictions such as overfitting, underfitting, passiveness towards spatial variances [7], limited feature extraction capabilities [8] and neural saturation [9]. These training afflictions lead to premature convergences to local optima rather than a true convergence to the global optimum. In this paper, we focus on saturation in the convolutional block of a CNN model as it is relatively less explored compared to the other training afflictions previously mentioned. Saturation occurs predominantly during the backpropagation step of learning in neural networks when a non-linear differentiable activation function $\rho$ is applied to an input dimensional vector $\mathbf{x} = \langle x^1, x^2 \cdots x^d\rangle$. A neuron is said to be fully saturated when its output $y = \rho(\sum_{i=1}^{n} \omega_i x_i + b_i)$, acquires values close to the bounded subset of $\rho$, where $\rho$ is the non-linear activation function, $\omega_i$ is the computed convolutional filters from which a regression vector is calculated, $b_i$ is the bias, $\boldsymbol{x_i} \in \boldsymbol{X}$. In order for a neuron to be fully saturated, the gradient($\partial$) of $y$ relative to the input $\boldsymbol{x_i}$, $\partial y/\partial x_i$ approaches zero for a logistical activation function. Inversely, the weights associated with such fully saturated neurons must be substantial to reach the bounded subset of $\rho$. In other words, a neuron is saturated when the weight updates calculated from the error are so negligible that it causes no apparent change in the magnitude affecting convergence.

While there has been prior research on neural saturation in ANNs, including proposed methods to mitigate this phenomenon [9], to the best of our knowledge there is no literature studying this same concept in the convolutional block of a CNN. In this paper, we focus on understanding kernel saturation during the generation of feature maps in CNNs and propose a novel data augmentation technique for mitigating saturation of convolutional kernels. We hypothesize that supplementing training data with negative images will lead to more efficient extraction of semantic feature information, partially offsetting saturation of convolutional kernels, detailed extensively in Section 6-B. To generate negative images, a logical bitwise *NOT* operations are used, as it does not significantly modify the structural information of images in dataset, explained more in Section IV-A. Furthermore, the logical bitwise *NOT* operation used to create negative images can be performed in real-time using the less computationally intensive CPU rather than the GPU during model training, reducing computational overheads.

We test our research hypothesis using the RESidual NETwork (ResNet) CNN architecture [6] on two well known benchmarking datasets, CIFAR-10 [10] and STL-10 [11]. The ResNet architecture is chosen as it introduces stacked convolutional layers and skip connections which ensure only the most important feature information contributing to classification performance is retained. The CIFAR-10 and STL-10 datasets were chosen since they have similar image classes but differ in image resolutions thus making for a better understanding of the saturation phenomenon. We also employ two quantitative measures, Maximum Entropy (ME) and Signal to Noise Ratio (SNR) to analyze the informational content present in the datasets.

Our results show that there is a statistically significant increase in classification accuracy when negative images are supplemented to the original datasets even though the ME measures do not increase significantly. Our reasoning for this increase in accuracy is that when the image data from the supplemented dataset is normalized, only the relevant semantic information is used for kernel weight updates while random structural differences in the images caused by noise become irrelevant, discussed more in Section 6. Furthermore, convolutional kernel saturation can be offset by using negative images as the loss function must effectively alternate between maximization and minimization as the images are essentially inverses to one another. This alternation encourages more controlled weight updates and is a more effective backpropagation mechanism.

The main contributions of this paper are, presenting evidence to show CNNs are indeed susceptible to convolutional kernel saturation, and present a novel data augmentation technique that mitigates kernel saturation. The background information needed for the contributions made is outlined in Section 2, the theoretical basis for kernel saturation and the mathematical reasoning for showcasing the effectiveness of augmenting datasets with negative images is presented in Section 3. The experimental design and empirical validation are explained in Sections 4 and 5. Finally, discussion of the results, limitations of the study and the conclusion are highlighted in Sections 6, 7-A and 7.

## 2. BACKGROUND

The accuracy of any Neural Network (NN) depends on generating adequate internal feature representations from the input data. In CNNs, the feature representations are separated into $N$ feature maps generated from the input image of pixel size $R \times R$ using $N$ convolutional kernels with weight matrices $\omega$. The feature maps are hierarchical, meaning the feature maps generated at a given layer $L$ are computed from the data propagated from the preceding layer $L-1$. The three-dimensional input color images ($2D$ image of size $R \times R$ with 3 color channels) are dimensionally reduced down to $N$ $2D$ feature maps of size $S \times S$, such that $R \geq S$. Dimensionally reducing a higher-order non-linearly separable complex function into a set of lower-order linearly separable matrices is not always exact and may exclude critical information. This limitation can be mitigated by introducing a non-linear activation function $\rho$ [12].

Assuming the input image is fed at the input layer with pixel $R_{0,0}$ indexed by $R_{i,j}^0$, the convolutional output at the $n^{th}$ layer is denoted by $y_i^n$. The backpropagation error can be calculated using the $n^{th}$ layer gradient $\partial y_0^n / \partial R_{0,0}^{n-1}$ and the loss function $l$ as $\partial l / \partial y_0^n$. Applying the chain rule to the error for a lower dimensional output $d'$ we get $\partial l / \partial R_{0,0}^{n-1} = \sum_{i'}^{d'} [(\partial l / \partial y_{i'})(\partial y_{i'} / \partial R_{i'}^{n-1})]$ [13]. The sum of all the convolutional kernels/filters performing a linear transformation for $x_i$ for a given layer $L$ is equal to $2^N \Delta$ [8], where $\Delta = k \times k$ is the kernel width for the $N$ convolutional neurons/unit subsampled by 2.

**A. ResNet**

One of the distinguishing characteristics of the ResNet architecture is the non-uniform propagation of information implemented through skip connections. Skip connections facilitate shorter gradient propagation to initial layers, eliminating the problem of vanishing and exploding gradients. The error minimization function can be computed using the approximation vector $\langle \varphi(x_i) + x_i, \omega \rangle$. The output of a residual learning block when the gradient is small resolves into identity transformations, which has been shown to increase performance [6]. Even though a few of the training afflictions referred to in Section I are resolved by the ResNet architecture, true convergence is not achieved.

**B. Maximum Entropy in Image Data**

Entropy measures are widely used in image processing for image enhancements such as de-noising and image restoration/reconstruction using de-convolution [14][15]–[16]. Maximum Entropy (ME) is used to measure the maximum amount of information in images. The reason for using ME in this paper is to prove that the structural information in the negative images is equivalent to the standard images. The method of calculating ME [17] is given in equation 1

$$ME = \log_2(s^i) \; bits \tag{1}$$

Where, $i$ is the number of independent choices that can be made with $s$ number of distinct symbols. In grayscale images, $i$ would be 256 for the 0-255 gray levels with a 0 value for black and 255 for white and $s$ would be 784 for an image size of $28 \times 28$. ME measures are calculated separately for each of the color channels i.e. in case of color images Red, Green and Blue (RGB) and then averaged to get the final ME measure.

As an image is dependent on neighboring pixels to represent information the relative probabilities of each individual pixel are near impossible to calculate, the open-source scikit-image processing library written in python can be used to calculate the ME measures for color and grayscale images. SciKit-image processing library uses a disk (set to the size of the input training image) to scan across the input data and return the frequency count of color levels. Using this method we can calculate the maximum entropy measures for CIFAR-10 and STL-10 datasets as **6.850, 6.907** bits and **6.850, 6.907** for their negative counterparts respectively.

### C. Signal and Noise in Image Data

Accurate, quantifiable estimation of image quality, regardless of variance, plays a pivotal role in applications of digital image processing. There are many measures to mathematically calculate digital image quality, including Mean Squared Error (MSE), Root Mean Squared Error (RMSE), Signal-to-Noise Ratio (SNR) and Peak Signal-to-Noise Ratio (PSNR) [18]. According to [18], measures that consider the human visual system that integrate perceptual quality measures offer no distinct advantages over existing methods such as PSNR. The advantage of SNR over PSNR is that the former is the average intensity rather than the maximum squared intensity rather of the latter.

PSNR would be a less strict criterion to use for accurate signal measurement compared to SNR since PSNR $\geq$ SNR and PSNR would be significantly affected for constant signals whose variances are null but power variances are not null. The definition of SNR varies depending on the research field [19] but according to [20], SNR is a measure that compares the level of the desired signal to the level of background noise in the fields of science and engineering. In this paper, the use of SNR is warranted since we aim to understand kernel saturation and the effect of weight updates through supplementing negative images, which in theory would have an inverse SNR to the standard images. Mathematically, SNR in digital images is defined as the ratio of the quotient of mean signal intensity to the standard deviation of the noise [21] and is given by Equation 2.

$$SNR = \mu(\bar{S})/\sigma_N \tag{2}$$

Where, SNR is the signal to noise ratio (unit-less), $\mu(\bar{S})$ the mean of signal data and $\sigma_N$ is the standard deviation of the signal data with respect to the random noise.

The equation for calculating the mean of signal data is given in Equation 3,

$$\mu(\bar{S}) = (\Sigma_{i=1}^n S)/n \tag{3}$$

Where, $S \in [0-255]$, $\mu(\bar{S})$ is the mean of signal data when the pixel values for $S$ are in between 0-255 for red, green and blue color channels for CIFAR-10 and STL-10.

The equation for calculating standard deviation of signal data with respect to noise is given by Equation 4.

$$\sigma_N = \sqrt{(1/n)\Sigma_{i=1}^R (R_i - \mu(\bar{S})^2)} \tag{4}$$

Where, $S \in [0-255]$, $\sigma_N$ is the standard deviation of the data, i.e. the signal data with respect to the noise, $\mu(\bar{S})$ is the mean of signal data calculated using equation 3. $R$ is the total number of pixels in the data, $R_i$ is the value of the $i^{th}$ pixel in the flattened image.

Using equation 2, we can calculate SNR for CIFAR-10 and STL-10 with the mean of signal data calculated using Equation 3 and standard deviation computed using equation 4. The SNR values for CIFAR-10 and STL-10 are **2.39, 1.99** and for their negative versions are **2.73, 2.66** respectively.

## 3. ME, SNR IN QUANTIFYING INFORMATION PROPAGATION IN CNNS

ME in digital image processing provides the greatest amount of usable information in the image [15], while SNR indicates the quality of signal information for the image [22]. In other words, SNR reflects the extent to which the signal information is corrupted by random noise. ME and SNR measures are critical in quantitatively examining the input data for analyzing the effectiveness of kernel weight updates by measuring CNN models classification performance. This is because larger ME measures indicate greater usable signal information in the data, which warrants higher abstractions. Larger SNR values indicate the extent to which this signal/feature information can be extracted, which also corresponds to the extent of complex feature map abstraction. If CNNs are susceptible to kernel saturation, our hypothesis is that supplementing the training data with negative images having the same structural components (indicated through their ME measures), but inverse SNR information, might yield more effective weight updates due to the inherent spatial information but varied representations. Weight updates are performed using the backpropagation of errors using a partial derivative of the neural output computed using the gradient descent approach for error minimization of a loss function $l$. The loss function is calculated using the difference between the initial random kernel weights $W$ and the output $y$ for a given input $x_i$, i.e. $l(W, y(x_i) = (W - y)^2$ for a Mean Squared Error function.

The kernel weight update in CNNs depends on the amount of information (ME) in conjunction with the quality and complexity of signal information (SNR) present in the images. A low SNR indicates that there is more noise than signal information and low ME measures imply that there is less usable signal information in the images compared to noise. In circumstances where images have low SNRs like those present in the MNIST dataset, any attempts made to recover the original signal information using inverse filtering and other such methods produce convolutional outputs of unacceptable quality. The quality is reduced because, according to [22], noise and signal are intertwined implying that noise in the data introduces distortions and errors, which leads to uncertainties. These uncertainties introduce a more considerable weight change than is warranted. The classification accuracy for the unaugmented MNIST dataset is already high (99.41%), since the signal data is simple and can be linearly separated. Substantial weight changes for simple datasets do not affect neural outputs since the global optimum can be easily found and therefore in these scenarios negative image data augmentation offer no significant increases in classification accuracies.

Assume an image has a ME of 1, this would imply that to fully reconstruct the image, a single bit of information is sufficient. Now assume the SNR measure is 1, suggesting there is no corruption of the signal with noise. Therefore for complete classification accuracy, a single bit of information is sufficient for backpropagation and weight adjustment i.e. if the CNN outputs 0, the error will be 100% and the weight update enables the updated CNN to output 1. However, if the signal is corrupted by noise, the weight updates and error calculations need to be adjusted such that the noise or random variances are taken into account. Say the ME remains at 1, but SNR is 0.5, then the weight updates are calculated using the partial derivative of the loss explained earlier in this section.

## 4. EXPERIMENTAL DESIGN

Experimentation revolves around testing the effect of different ME and SNR measures on the saturation of convolutional kernels and the effect of saturation on convergence to the global optimum in deep CNNs. A quantitative methodology was employed to collect the test-set classification accuracy for the three datasets, MNIST [23], CIFAR-10 [10], STL-10 [11]. The research hypothesis is that supplementing negative images increases classification accuracies. The hypothesis was examined against published standard state-of-the-art ResNet models, while keeping all other HPs such as learning rate and batch size constant with no data excluded or pre-processing steps applied to images in the datasets for three evaluation runs of 500 epoch instances. Learning rate was set to 0.001 using an adaptive optimizer (ADAM) and a batch size of 128 was selected based on configurations by the original authors of the proposed architectural models [6].

### A. Datasets
The CIFAR-10 dataset consists of 32 ×32 color natural images with ten classes including cars, airplanes and dogs. The dataset is balanced across 50,000 training and 10,000 testing images. The STL-10 dataset includes 500 training and 800 test natural color images split into much of the same classes of natural images but in a higher 96× 96 resolution, derived from the ImageNet dataset. The standardized testing protocol for STL-10 is not adopted in this paper, as the aim is to evaluate model accuracy for supervised learning problems instead of unsupervised feature extraction. To obtain the negative images of all the samples in the datasets we perform logical bitwise *NOT* operations on every pixel separately on the three color channels, Red (R), Green (G) and Blue (B) using the OpenCV python library.

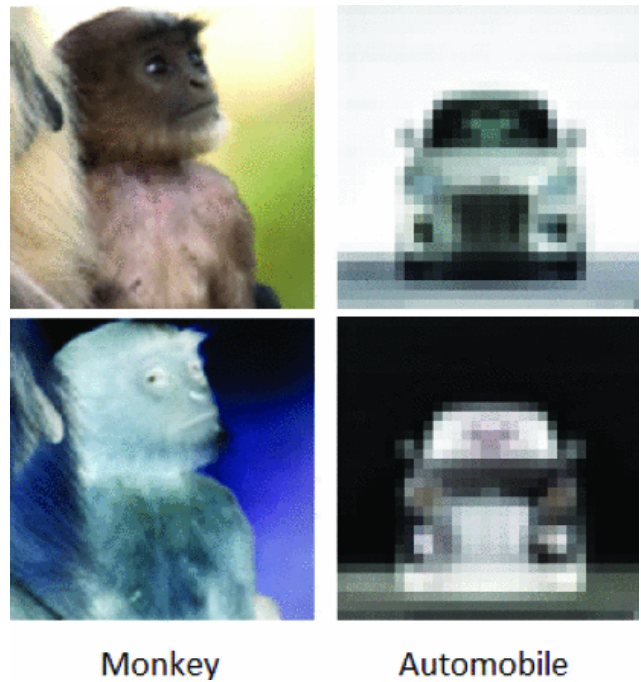

Fig. 1. Illustration of random negative sample images for two class in the STL-10 and CIFAR-10 datasets

### B. Experimental Setup
All experimentation was performed using a single NVIDIA 2080ti and Tesla P100 GPU with 12GB and 16GB of VRAM generously provided by InfuseAI Limited and the New-Zealand e-Science Infrastructure (NeSI) respectively. All models were saved after the predetermined number of epochs and each model was tested on the standard test-set images. The summary Table I presents the averaged results for the three 500 epoch runs. The training-validation split for all models was kept constant at 80%-20% for all the datasets, batch-size of 128 was based on the original ResNet paper [6], learning rate of 0.001 and an adaptive optimizer (ADAM) was used for faster convergence. There were no modifications made to the ResNet architecture with no image augmentation methods used to ensure reproducibility.

## 5. RESULTS

The averaged experimental results are presented in Table I for the two standard datasets along with the same datasets supplemented with negative images and training using only the negative images. Models were trained using only the negative images for 500 epochs only once. Further repetition was deemed unnecessary as there were significant differences that can realistically only arise when the models are not sufficiently converging to the input data.

### A. Statistical Analysis
First, the Shapiro-Wilk test for normality was used to establish if the raw data was normally distributed. The data was normally distributed with all p-values meeting the 5% threshold. To discard any interpretations of the results which might be due to random chance, we select the

Table I. SUMMARY TABLE OF RESULTS WITHOUT PRE-PROCESSING OR REAL-TIME DATA AUGMENTATIONS

| CNN Model | Dataset | Accuracy |
|---|---|---|
| **Standard datasets** | | |
| ResNet-50 | MNIST | **99.41 %** |
| ResNet-50 | CIFAR-10 | 80.65 % |
| ResNet-50 | STL-10 | 56.08 % |
| **Datasets augmented with negative images** | | |
| ResNet-50 | MNIST | 99.35 % |
| ResNet-50 | CIFAR-10 | **83.81 %** |
| ResNet-50 | STL-10 | **63.06 %** |
| **Datasets with only negative images** | | |
| ResNet-50 | MNIST | 33.95 % |
| ResNet-50 | CIFAR-10 | 43.73 % |
| ResNet-50 | STL-10 | 27.96 % |

The standard test-set was used for measuring performance parametric paired t-test for statistical testing of the data. A paired t-test is the most applicable since we want to question if there is an observable difference in accuracies for the two sets of data, the standard dataset and the supplemented standard dataset on the same architectural model. In other words, is there a statistical difference in the classification accuracies when standard datasets are supplemented with negative images?

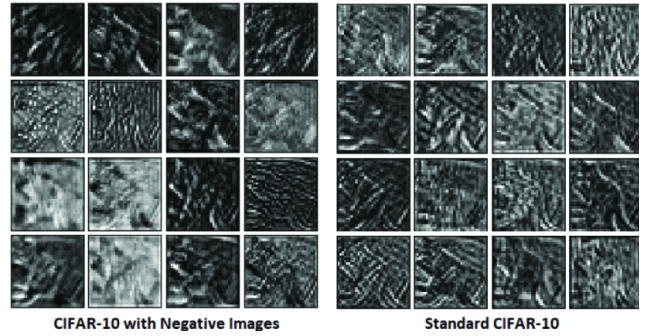

Fig. 3. Visualized activation maps after the final convolutional layer for a trained ResNet-50 architecture (89 epochs)

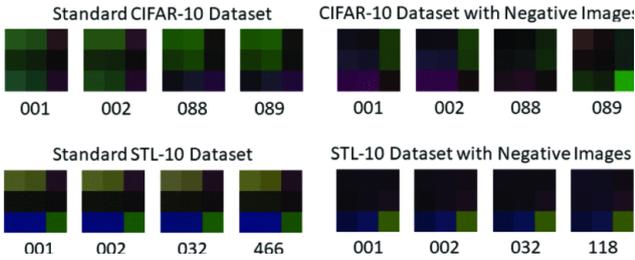

Fig. 2. Visualized convolutional kernel weights with epoch instances for two experimental datasets extracted from a trained ResNet-50 architecture

Analysis was performed with the independent variable being the input dataset and the dependent variable being classification accuracy. Interpretation was done at the standard significance cut-off level of 0.05 using a two-tailed test. A two-tailed test is warranted as the assumption is that the standard dataset supplemented with negative images might provide a higher classification accuracy compared to standard dataset but the inverse is also possible. The default null hypothesis is that no observable differences in accuracy are present. Both the natural image datasets (CIFAR-10 and STL-10) on the ResNet-50 showed statistical significance with p-values of 0.0013 and 0.0158 with means and variances of (80.65, 83.81), (56.08, 63.06) and (0.0121, 0.0157), (3.5087, 0.1905) respectively.

## 6. DISCUSSION

### A. Overview

A visual representation of the kernel weights for the first convolutional layer of the ResNet-50 CNN architecture is presented in Figure 2 for the two experimental datasets and the same datasets supplemented with negative images. The visualizations are generated across multiple epoch instances for the initial, middle and last epoch where the model is improved from previous instances. Figure 2 illustrates the kernel weight adjustments for the CIFAR-10 dataset are more varied and spread out across a larger set of epochs. Higher variance in the kernel weight updates can also be observed across the STL-10 dataset. The negative images are quite indistinguishable to the standard pictures, the ME and SNR measures for the negative images are 3.452 and 3.02, respectively. The standard dataset has ME and SNR measures of 3.139 and 0.44 respectively, which implies the weight changes are highly tuned towards converging on a single set of images with the others being neglected as they offer little to no significance in the error backpropagation. In other words, the global optimum can be converged to easily and therefore adding negative images offers no significant improvement in classification accuracy, unlike complex natural images which behave differently.

We can clearly see from Figure 3 that the feature extraction is smoother and more controlled for the dataset supplemented with negative images compared to the standard dataset. Furthermore, there are no adverse impacts from supplementing negative images. The ME values for the supplemented datasets do not significantly increase, as the amount of semantic information in the datasets remains the same. This is in stark contrast to the SNR measures which merely indicates that more number of pixels are brighter thus negative images have higher SNR measures. The difference in the SNR measures for the two datasets, however small, indicates the true signal variance $\sigma_s$ which forces kernel weight updates close to the mean ignoring pseudo-random variances caused by random noise. In other words, the loss function for a weight matrix $\omega_{x_i}$ using $\mathbf{x}_i$ feature inputs is $l(\omega_{x_i})$.

### B. Reasoning

Assuming that the elements of $\omega_{x_i}$ are represented in only a 2-dimensional space then the loss function $l\omega_{x_i}$ has an easy interpretation in a 3-dimensional space which can be trivially optimized for linear separation. Problems occur when the loss function remains unchanged over a large number of epochs making future kernel weight change susceptible and negligible weight updates cause premature convergences to local optima. This is because the backpropagation error calculated using the $n^{th}$ layer gradient $\partial y_0^n / \partial R_{0,0}^{n-1}$ and the loss function $l$ as $\partial l / \partial y_0^n$ tends to ever-decreasing values for similar input thus saturating for any new input.

Now assume the loss function for a weight matrix $\omega_{x_i^{-1}}$ using $x_i^{-1}$ negative feature inputs is $l(\omega_{x_i^{-1}})$.

The $n^{th}$ layer gradient for this new loss function would have similar semantic information (indicated by the corresponding maximum entropy measure) but would require different feature representations (evidenced through varied SNR measure). Therefore, the back-propagation error would be equivalent to maximizing the previous loss function. In other words $l(\omega_{x_i}) = 1/l(\omega_{x_i^{-1}})$. The effect of this would be alternating between a maximization and minimization of the loss function.

### C. Special condition

The special condition witnessed from the experimental data is that, even though there is an apparent increase in the classification accuracy, supplementing negative images will not be effective if the model gets saddled in a local optimum with an architecture that offers no skip connections such as VGGNet. Since there are no skip connections in the VGGNet architecture, kernel weight updates become close to zero causing saturation from which the model cannot recover. In these instances, supplementing negative images might decrease the probability of saturation but will not eliminate kernel saturation. Although a simple data augmentation technique such as using negative images show apparent improvement in classification accuracy, guarantees in convergence or mitigating neural saturation cannot be made.

## 7. CONCLUSION

We have reported in this paper that convolutional kernels are indeed susceptible to saturation, causing premature local optima convergences. We have proposed a novel real-time image augmentation technique using bitwise logical *NOT* operations on datasets to mitigate convolutional kernel saturation and increase classification accuracy. The approach was empirically validated using the ResNet CNN architecture, ResNet-50 model on three well-known benchmarking datasets, MNIST, CIFAR-10 and STL-10. Our results show smoother and more controlled kernel weight updates offering 3-7% increase in classification accuracy on CIFAR-10 and STL-10 datasets when compared against CNN models trained on standard datasets.

### A. Limitations and Future work

Similar to other approaches used to enhance learning algorithms, augmenting/supplementing datasets with negative images does not necessarily yield perfect classification accuracies as seen through experimentation on the MNIST dataset. While the data augmentation technique has a solid mathematical basis for being implemented, convergence or mitigating neural saturation cannot be guaranteed. Furthermore, the statistical tests were performed with three data points and empirical validation of the technique was conducted on two datasets and one network architecture. Therefore, further experimentation using ImageNet and MS COCO datasets are reserved as future work along with examining the detrimental characteristics for simple datasets like MNIST. Our hypothesis is that negative images might offer increased generalization ability which requires further thorough experimentation reserved as future work.